\pdfoutput=1
\documentclass[11pt]{article}

\usepackage[]{ACL2023}

\usepackage{times}
\usepackage{latexsym}

\usepackage[T1]{fontenc}
\usepackage[utf8]{inputenc}

\usepackage{microtype}

\usepackage{inconsolata}

\usepackage{graphicx}
\usepackage{amsmath}
\usepackage{enumitem}
\usepackage{caption}
\usepackage{subcaption}
\usepackage{booktabs,arydshln}
\usepackage{url}

\title{CIKMar: A Dual-Encoder Approach to Prompt-Based Reranking in Educational Dialogue Systems}

\author{ Joanito Agili Lopo \and Marina Indah Prasasti \and Alma Permatasari \\
        Department of Computer Science and Engineering \\ Universitas Gadjah Mada \\
        \texttt{name@mail.ugm.ac.id}
        }
        
\begin{document}
\maketitle
\begin{abstract}
In this study, we introduce CIKMar\footnote{\url{https://github.com/joanitolopo/cikmar-system}}, an efficient approach to educational dialogue systems powered by the Gemma Language model. By leveraging a Dual-Encoder ranking system that incorporates both BERT and SBERT model, we have designed CIKMar to deliver highly relevant and accurate responses, even with the constraints of a smaller language model size. Our evaluation reveals that CIKMar achieves a robust recall and F1-score of 0.70 using BERTScore metrics. However, we have identified a significant challenge: the Dual-Encoder tends to prioritize theoretical responses over practical ones. These findings underscore the potential of compact and efficient models like Gemma in democratizing access to advanced educational AI systems, ensuring effective and contextually appropriate responses.
\end{abstract}

\section{Introduction}
The emergence of powerful Large Language Models (LLMs) such as ChatGPT has been proven effective in various tasks, including generating text that is nearly indistinguishable from human-written text \citep{KASNECI2023102274,omidvar-an-2023-empowering}. Building on the success in text generation, LLMs have shown significant potential in various applications, especially in the educational domain. 

In recent years, there have been various efforts to utilize these powerful large language models (LLMs) in education. They have been deployed in teacher-student collaborations as virtual tutors, guiding students through exercises, offering personalized learning experiences, and providing intelligent tutoring \citep{kamalov2023new}. Additionally, they are used for adaptive assessments and serve as conversational partners in learning scenarios \citep{,tan2023applying,li2024adapting}. 

Despite these promising opportunities, the use of generative models as a foundation for downstream tasks presents several crucial challenges such as inconsistently delivering accurate and contextually appropriate responses \cite{tack-etal-2023-bea}. Furthermore, Language models in current scenarios mostly use extremely large language models in terms of their parameter size, such as proprietary 175 and 137 billion-parameter GPT-3 model \citep{brown2020language}, or open source LLMs such as 70 billion-parameter LLaMA2 \citep{touvron2023llama}, 14 billion-parameter Qwen \citep{bai2023qwen}, and 6 billion-parameter ChatGLM3 models \citep{zeng2023glm130b}. Language models at this scale are not practical and inaccessible for many researchers and even practitioners, due to their large memory consumption and slow generation times \citep{ding2024efficiency,jimenez-gutierrez-etal-2022-thinking}, data privacy, and inflexibility of customization \citep{sinha2024evaluating}. Therefore, it is essential to determine how solid that foundation is and how it can be accessible for further development, especially in the educational domain. 

\begin{figure}[!t]
    \includegraphics[width=\linewidth,trim={0, 2em, 0, 0}, clip]{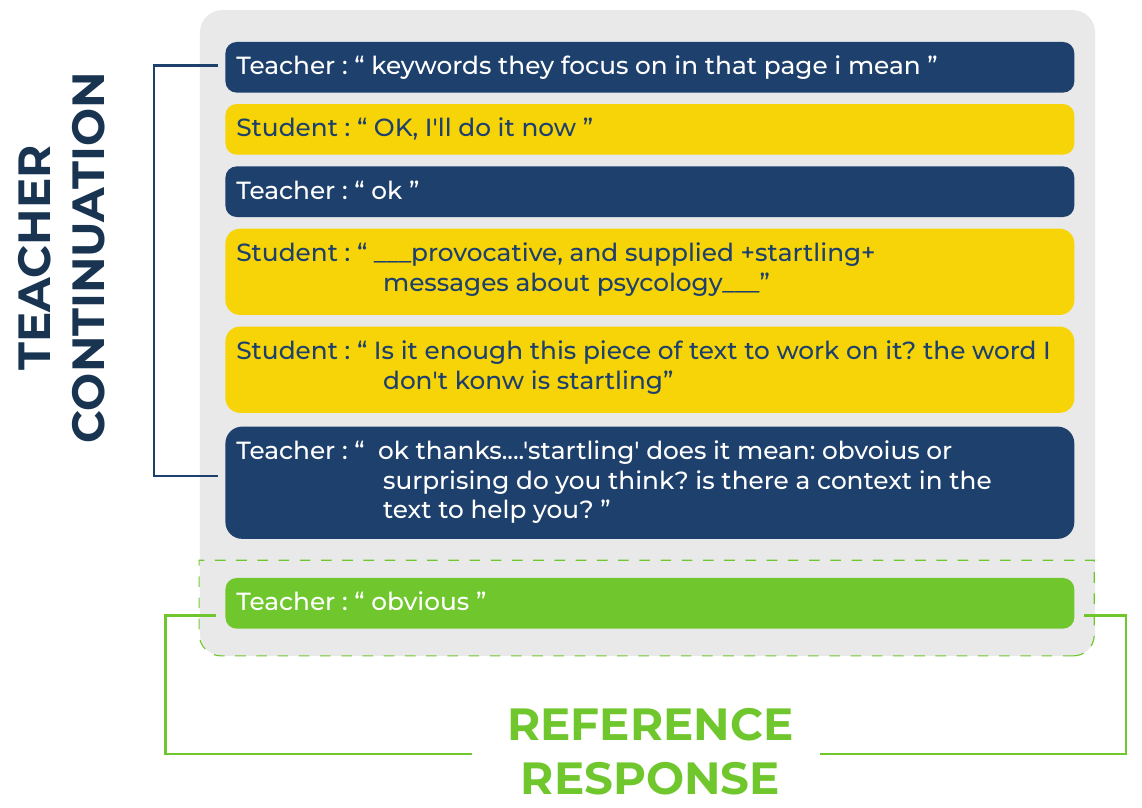}
    \caption{Teacher Continuation Data Visualization}
    \label{fig:teacher_continuation}
\end{figure}

According to the challenges above, we designed a simple but effective approach by leveraging Large Language Models and prompt-and-rerank approach \citep{suzgun-etal-2022-prompt} to build the dialogue AI system especially in educational domain. We chosen to work with a smaller, pre-trained language model called  Gemma 1.1 2B (IT), which can run efficiently on less than 12 GB of RAM and a single GPU T4. This makes it suitable for real-world applications by maintaining a reasonable model size without compromising performance. Additionally, a Dual-Encoder approach strategy has been adopted to rerank the candidate outputs generated by the model using hand-written prompts. This approach aims to increase the relevance and effectiveness of the responses generated by our system in educational dialogues.

\section{Related Work}
% Large Language Models (LLMs), like GPT-4, have proven effective in many domains but are not universally applied to all tasks \citep{naveed2024comprehensive}.  
Researchers have extensively investigated the effectiveness of various approaches utilizing language models. \citet{sridhar2023hierarchical} enhanced LLM performance on web navigation tasks using Actor-Summarizer Hierarchical (ASH) prompting, while \citet{kong2024better} improved reasoning benchmarks with role-play prompting. \citet{kojima2023large} showed that modifying prompt structure enables LLMs to perform multi-step reasoning in zero-shot settings.

In the educational context, \citet{adigwe-yuan-2023-adaio} and \citet{hicke-etal-2023-assessing} used GPT-3 and GPT-4 to generate educational dialogue responses, achieving high DialogRPT and BERTScore results with hand-written zero-shot prompts. Similarly, \citet{vasselli-etal-2023-naisteacher} used GPT-3.5 Turbo with manual few-shot prompts based on DialogRPT selection, which contributed most to the final outputs.

% Fine-tuning has also proven effective by utilizing large language models (LLMs) in educational domain. \citet{baladon-etal-2023-retuyt} applied the LoRa method to fine-tune larger models such as BLOOM-3B, Llama 7B \citep{touvron2023llama}, and OPT 2.7B \citep{zhang2022opt}. They found that even with the smaller parameter OPT 2.7B model, careful fine-tuning and processing could achieve better performance, offering both variety and considerable power. In line with these results, \citet{huber-etal-2022-ccqa} showed that using GPT-2 with 124 million parameters, enhanced with reinforcement learning through the NLPO algorithm \citep{ramamurthy2023reinforcement} for policy optimization during training, achieved high scores in terms of BERTScore.

% Due to the high computational power required for fine-tuning LLMs and the challenge of domain adaptive, \citet{omidvar-an-2023-empowering} introduced semantic in-context learning, which utilizes available private knowledge sources to provide accurate answers to inquiries. Furthermore, \citet{gu2024minillmknowledgedistillationlarge} proposed a method for reducing the size of LLMs through knowledge distillation, where a smaller student model is trained to replicate the performance of a larger teacher model. Their experiments with the distilled versions of GPT-3 showed that these compact models could perform competitively on various benchmarks. 

Fine-tuning has also proven effective by utilizing large language models (LLMs) in educational domain. \citet{baladon-etal-2023-retuyt} used the LoRa method to fine-tune models like BLOOM-3B, Llama 7B \citep{touvron2023llama}, and OPT 2.7B \citep{zhang2022opt}. They found that even the smaller OPT 2.7B model, with careful fine-tuning, could achieve better performance. Similarly, \citet{huber-etal-2022-ccqa} demonstrated that GPT-2, enhanced with reinforcement learning via the NLPO algorithm \citep{ramamurthy2023reinforcement}, achieved high BERTScores.

Due to the high computational power needed for fine-tuning and domain adaptation, \citet{omidvar-an-2023-empowering} introduced semantic in-context learning, using private knowledge sources for accurate answers. \citet{gu2024minillmknowledgedistillationlarge} proposed reducing LLM sizes through knowledge distillation, training smaller models to replicate larger ones. Their experiments with distilled GPT-3 versions showed competitive performance on various benchmarks.

Our research aims to develop an educational dialogue system using Gemma 1.1 IT 2B. This system uses prompts to guide LLMs in generating outputs based on contextual understanding, relevance, engagement, clarity, and feedback. To optimize results, it employs dual encoders (BERT and SBERT) to rerank top candidates. Our objective is to democratize open model LLM in real-world scenarios, ensuring accurate, relevant responses while enhancing student engagement and understanding in educational dialogues.

\section{Methods}

\begin{figure*}[t]
    \centering
    \includegraphics[width=\linewidth]{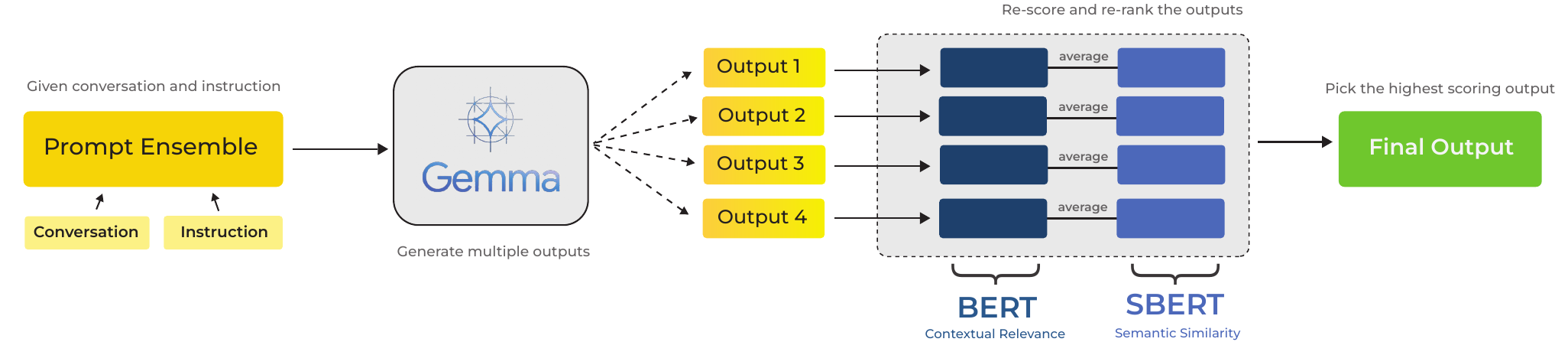}
    \caption{An illustration of the CIKMar system. Given an input conversation and instruction, we create the prompt ensemble and feed it to Gemma to generate multiple outputs. We then re-score each output by averaging BERT and SBERT scores and select the candidate with the highest re-ranked score as the final output}
    \label{fig:arsitektur}
    \vspace{-6pt}
\end{figure*}

\subsection{Data}
We used data from the BEA 2023 shared task, sourced from the Teacher-Student Chatroom Corpus (TSCC) \citep{caines-etal-2020-teacher2, caines-etal-2022-teacher}. This corpus consists of several conversations where an English teacher interacts with a student to work on language exercises and assess the student's English proficiency \citep{tack-etal-2023-bea}. Each conversation contains multiple responses and starts with either \textbf{teacher:} or \textbf{student:} prefixed. The reference text is the teacher's response that follows the previous input dialogue. The corpus includes a training set of 2,747 conversations, a development set of 305 conversations, and a test set of 273 conversations, totaling 3,325 conversations.

Since the data was collected from real-time teacher-student conversations, turn-taking is not as consistent as in most dialogue systems \citep{vasselli-etal-2023-naisteacher}. There are two patterns mostly occur: conversations ending with the student (teacher reply) and conversations ending with the teacher (teacher continuation). This condition occurs in 38\% of the training data and 40\% of the development data. Figure~\ref{fig:teacher_continuation} shows an example of a conversation in the teacher continuation condition.

\subsection{Prompt Ensemble}
We utilized hand-written prompts from \citet{vasselli-etal-2023-naisteacher} to build our system. The prompts include Zero-shot and Few-shot types, targeting both general and specific scenarios. We used only the five main prompts available as they are already tailored for teacher responses and continuations. This selection also ensures general applicability to other datasets or conversations. For full details explanation of each prompt take a look at Appendix~\ref{sec:prompts_appendix}. 

In the creation of the few-shot prompts, it requires positive and negative examples to help the model avoid irrelevant responses. We adopt the method of \citet{vasselli-etal-2023-naisteacher} who applied the handcrafted, generative, and iterative prompt methods. However, we modified the iterative method from the original paper. Instead of using DialogRPT, we employed the BM25 ranking function to select the highest and lowest scoring responses as positive and negative examples.

BM25 \citep{10.1007/978-1-4471-2099-5_24,10.1561/1500000019} was chosen over DialogRPT because it reduces the computational power required for the prompting and re-ranking process, as DialogRPT needs additional memory capacity to calculate and choose the best candidate. Additionally, BM25 is known as the first-stage ranker in lexical retrieval systems \citep{askari2023injecting} which ensures positive and negative examples are selected based on their lexical match with the conversation history.

\subsection{Gemma Instruct-tuned Model}
Our main system leverages a pretrained language model with a prompting approach rather than training one from scratch or fine-tuning it on a new dataset. We used the Gemma 1.1 IT 2B model \citep{gemmateam2024gemma}, 2-billion parameter open model developed by Google for efficient CPU and on-device applications. The model has shown strong performance across academic benchmarks for language understanding, reasoning, and safety, such as MMLU \citep{hendrycks2021measuring}, SIQA \citep{sap-etal-2019-social2}, HumanEval \citep{chen2021evaluating}, and Winogrande \citep{sakaguchi2019winogrande}. These results indicate its promising performance in educational contexts.

\begin{table}[ht!]
    \setlength{\tabcolsep}{6pt}
    \centering
    \footnotesize   
    \begin{tabular}{r l}
    \toprule
    \vspace{0.2cm}
    \textbf{User:} & {\color{blue}\texttt{<start\_of\_turn>user}} \vspace{-0.2cm} \\
    & \texttt{conversation} \\
    & \texttt{instruction}
    {\color{blue}\texttt{<end\_of\_turn>}} \\
    & {\color{blue}\texttt{<start\_of\_turn>model}} \vspace{0.1cm} \\
    
    \textbf{Model:} & \texttt{responses}{\color{blue}\texttt{<end\_of\_turn>}} \vspace{0.1cm} \\

    \bottomrule
    \end{tabular}
    \caption{Example dialogue with user and model control tokens.}
    \label{tab:sample_dialogue}
    % \vspace{-0.5cm}
\end{table}

We followed the instruction-formatted control tokens suggested in the Gemma technical report to avoid out-of-distribution and poor generation. Table \ref{tab:sample_dialogue} shows an example dialogue with user and model control tokens. Specifically, the relevant token \texttt{user} represents the role, and its \texttt{content} includes the conversation history followed by the prompt. Meanwhile, the \texttt{model} turn responds to the \texttt{user} dialogue.

In our experiments with the training and development sets, the Gemma model sometimes generated hallucinations on the first attempt, such as factually incorrect response, nonsensical content, overly long response, and content disconnected from the input prompt. However, performance improved on the second and third attempts. Therefore, to ensure the best response, we generated each candidate three times before selecting the final output.

We configured several parameters to control the model's output such as set the \texttt{max\_length} of the generated output to 512 tokens, \texttt{no\_repeat\_ngram\_size} to 2 to avoid repetition, and used \texttt{top\_k=50} and \texttt{top\_p=0.95} to balance randomness and coherence. The \texttt{temperature} was set to 0.7 for more conservative choices. Finally, we enabled probabilistic sampling over greedy decoding.

% Based on our experiment with training and development sets, Gemma 1.1 IT 2B models sometimes give hallucinate in first attempt of generation, such as the factually incorrect response, nonsensical, too long responses, and disconnected from the input prompt. Meanwhile, when we tried second and third attempts it gets better. Thus, the generation of each candidates we do it three times before take it as final output. In addition, we also set the \texttt{max\_length} of the generated output equal to 512 tokens for each generation.

% We configured several parameters to control the model's output. We set \texttt{no\_repeat\_ngram\_size} to 2 to avoid repetition, and used \texttt{top\_k=50} and \texttt{top\_p=0.95} to balance randomness and coherence. The \texttt{temperature} was set to 0.7 for more conservative choices. Finally, \texttt{do\_sample=True} enabled probabilistic sampling over greedy decoding.

\subsection{Dual-Encoder Reranking}
% Inspired by previous research \cite{vasselli-etal-2023-naisteacher,suzgun-etal-2022-prompt,haroutunian-etal-2023-reranking}, our system generates multiple candidates outputs from different manually designed prompts; then rerank the outputs by a heuristically defined scoring function. Specifically, for the scoring function we utilized S-BERT \citep{reimers-gurevych-2019-sentence2} and BERT \citep{devlin-etal-2019-bert2} by averaging cosine similarity scores from their embedding to find the fine-grained semantic relevance, context-response matching in embedding space between the conversation history and the generated response. 

Inspired by previous research \cite{vasselli-etal-2023-naisteacher,suzgun-etal-2022-prompt,haroutunian-etal-2023-reranking}, our system generates multiple candidate outputs from different manually designed prompts and then reranks these outputs using a heuristically defined scoring function. Specifically, for the scoring function, we utilize SBERT \citep{reimers-gurevych-2019-sentence2} and BERT \citep{devlin-etal-2019-bert2}, averaging cosine similarity scores from their embeddings to assess fine-grained semantic relevance and context-response matching in the embedding space between the conversation history and the generated response.

In the given setup, we start with a dialog as a context \texttt{ctx} and a list of candidate responses $\{\texttt{cand}_1, \texttt{cand}_2, \ldots, \texttt{cand}_m \}$. Initially, we compute SBERT and BERT embeddings for both the context and the candidate responses. For BERT embeddings we calculated by averaging token embeddings across the sequence dimension.

% \[ e_{\text{bert}} = \frac{1}{N} \sum_{i=1}^{N} H_i\]
% where $H$ is a matrix of size $(N, d)$ with $N$ being the number of tokens and $d$ the embedding dimension. The embeddings for the context and the $i$-th candidate response are denoted as $\ e^{\text{bert}}_{\text{ctx}}$ and $\ e^{\text{bert}}_{\text{cand\_i}}$, respectively. 

The cosine similarity between the context and each candidate response embedding, for both SBERT and BERT, is calculated using:
\[ S_{\text{emb}}(i) = \cos\left(e_{\text{ctx}}^{\text{emb}}, e_{\text{cand}_i}^{\text{emb}}\right) = \frac{e_{\text{ctx}}^{\text{emb}} \cdot e_{\text{cand}_i}^{\text{emb}}}{\|e_{\text{ctx}}^{\text{emb}}\| \|e_{\text{cand}_i}^{\text{emb}}\|}
 \]
where $\texttt{emb} \in \{\texttt{sbert}, \texttt{bert}\}$.

To combine these similarity scores for each candidate response, we average the SBERT and BERT similarity scores.
% \[
% S(i) = \frac{S_{\text{sbert}}(i) + S_{\text{bert}}(i)}{2}, \quad \forall i \in \{1, 2, \ldots, m\}
% \]

Finally, the candidates are ranked based on these combined similarity scores in descending scores. The indices of the candidates are sorted according to their combined scores, and it returns the list of candidates responses ordered from most to least relevant to the given context. 
 
% Next, BERT embedding are generated for the context and each candidate response by averaging token embedding across the sequence dimension. The $e_{\text{bert}}$ represents the BERT embedding for the context, computed as:
% \[ e_{\text{bert}} = \frac{1}{N} \sum_{i=1}^{N} H_i\]
% where $H$ is a matrix of size $(N, d)$ with $N$ being the number of tokens and $d$ the embedding dimension. Similarly, $\ e^{\text{bert}}_{\text{cand\_i}} $ denotes the BERT embedding of the $i$-th candidate response. The cosine similarity for BERT embeddings is then computed as:
% \[ S_{\text{bert}}(i) = \cos\left(e_{\text{ctx}}^{\text{bert}}, e_{\text{cand}_i}^{\text{bert}}\right) = \frac{e_{\text{ctx}}^{\text{bert}} \cdot e_{\text{cand}_i}^{\text{bert}}}{\|e_{\text{ctx}}^{\text{bert}}\| \|e_{\text{cand}_i}^{\text{bert}}\|}
%  \]

% To combine these similarity scores for each candidate response, we average the SBERT and BERT similarity scores: 
% \[
% S(i) = \frac{S_{\text{sbert}}(i) + S_{\text{bert}}(i)}{2}, \quad \forall i \in \{1, 2, \ldots, m\}
% \]

% Finally, the candidates are ranked based on these combined similarity scores in descending scores. The indices of the candidates are sorted according to their combined scores, and it returns the list of candidates responses ordered from most to least relevant to the given context. 

\subsection{Post-processing}
% The raw outputs of Gemma 1.1 IT 2B contained inconsistent formatting, sometimes including phrases prefixed by ``**'' or starting with unwanted preamble text, such as \texttt{Teacher:} or \texttt{Student:}. The model also produce the explanation of its response start with the \texttt{Explanation: } prefix, which makes it long unnecessary output. However, we found the pattern that the response always start with the \texttt{""}. To address this, post-processing was necessary to clean and standardize the responses. We defined a regular expression pattern, \texttt{\textbackslash*\textbackslash*.\textbackslash*?:\textbackslash*\textbackslash*\textbackslash n\textbackslash n}, to identify and remove any unwanted initial phrases. Each response was then processed to eliminate these phrases, ensuring a cleaner format. Furthermore, we process the first occurrence of a quotation mark in each response. If a quotation mark was found, only the text following it was retained, thus discarding any leading irrelevant content. Finally, any leading or trailing whitespace was removed from the cleaned text. This post-processing step was crucial for ensuring that the outputs from model were consistently formatted and more straightforward to utilize.

The raw outputs from model often included inconsistent formatting, such as phrases prefixed by "**" or starting with unwanted text like \texttt{Teacher:} or \texttt{Student:}. Additionally, the model sometimes appended lengthy explanations to its responses beginning with \texttt{Explanation:}, adding unnecessary length. However, we observed a consistent pattern where the actual response always began with a quotation mark \texttt{"}.

To standardize these outputs, we implemented a post-processing step. First, we defined a regular expression pattern, \texttt{\textbackslash*\textbackslash*.\textbackslash*?:\textbackslash*\textbackslash*\textbackslash n\textbackslash n}, to identify and remove any unwanted initial phrases. This pattern effectively removed prefixes like "**", \texttt{Teacher:}, or \texttt{Student:}. Next, each response was processed to retain only the text following the first occurrence of a quotation mark, discarding any preamble or unnecessary content. Finally, we trimmed any leading or trailing whitespace. 

% Next, each response underwent further processing to retain only the text following the first occurrence of a quotation mark. This ensured that any preamble or unnecessary content preceding the actual response was discarded.

% Finally, we trimmed any leading or trailing whitespace from the cleaned text. This post-processing was essential to ensure consistency in the format of the model outputs, making them more straightforward and reliable for use in subsequent analysis or applications.

% berisi detail dataset, penjelasan dari metode/model arsitektur yang digunakan, scenario eksperimen yang dilakukan (jika ada)

\section{Result \& Analysis}

\begin{table}[t]
    \centering
    \resizebox{\linewidth}{!}{
    \begin{tabular}{llll}
    \toprule
    \textbf{\#} & \textbf{Precision} & \textbf{Recall} & \textbf{F1-Score} \\
    \toprule
    CIKMar (\textbf{ours}) & 0.69 & 0.70 & 0.70 \\
    \midrule
    NAISTeacher \citet{vasselli-etal-2023-naisteacher} & 0.71 & 0.71 & 0.71 \\
    Adaio \citet{adigwe-yuan-2023-adaio} & 0.72 & 0.69 & 0.71 \\
    GPT-4 \citet{hicke-etal-2023-assessing} & 0.71 & 0.69 & 0.70 \\
    S-ICL \citet{omidvar-an-2023-empowering} & 0.72 & 0.69 & 0.70 \\
    OPT-2.7B \citet{baladon-etal-2023-retuyt} & 0.74 & 0.68 & 0.71 \\
    NLP-HSG \citet{huber-etal-2022-ccqa} & 0.72 & 0.63 & 0.67 \\
    Alpaca \citet{baladon-etal-2023-retuyt} & 0.72 & 0.68 & 0.70 \\
    DT \citet{tack-etal-2023-bea} & 0.67 & 0.62 & 0.64 \\
    % \midrule
    % Public datasheet & Datasheet review & 1 & 1 \\
    % Public datasheet & Dataloader review & 2 & 4 if difficult \\
    % Public datasheet & Private datasheet review & 0.5 & - \\
    % Public datasheet & Private data contact & 1 & 5 if succeeds \\
    \bottomrule
    \end{tabular}
    }
    \caption{Comparison of our proposed system with previous research based on BERTScore \citep{zhang2020bertscore}}
    \label{tab:main-results}
\end{table}

% \begin{figure*}[t]
%     \centering
%     \includegraphics[width=0.8\linewidth]{figures/2d_visualization.pdf}
%     \caption{Mapping between tasks, schemas, modalities, and language regions across 498 datasheets in SEACrowd.}
%     \label{fig:2d_visualization}
%     \vspace{-6pt}
% \end{figure*}

% \begin{figure*}[t]
%     \centering
%     \includegraphics[width=0.8\linewidth]{figures/similarity_heatmap.pdf}
%     \caption{Mapping between tasks, schemas, modalities, and language regions across 498 datasheets in SEACrowd.}
%     \label{fig:similarity_heatmap}
%     \vspace{-6pt}
% \end{figure*}

% \begin{figure*}[!t]
%   \centering
%   \begin{subfigure}[t]{0.45\linewidth}
%       \includegraphics[trim={0, 0, 0, 0}, clip, width=\linewidth]{figures/2d_visualization.pdf}
%       \caption{Language coverage}
%       \label{fig:resource-gaps-coverage}
%   \end{subfigure}
%   \begin{subfigure}[t]{0.45\linewidth}
%       \includegraphics[trim={0, 0, 0, 0}, clip, width=\linewidth]{figures/2d_visualization.pdf}
%       \caption{Annotation quality}
%       \label{fig:resource-gaps-quality}
%   \end{subfigure}
%   \caption{The resource gap in SEA in terms of language coverage, annotation quality, and cultural relevance.}
  
%   % \caption{SEACrowd covers 980 languages in Southeast Asia. There are $\sim$300 languages uncovered Southeast Asian languages remain after SEACrowd.}
%   \label{fig:resource-gaps}
%   \vspace{-8pt}
% \end{figure*}

\subsection{Main Result}

% \begin{table*}[!ht]
% \small 
% \centering
% \scalebox{1.0}{
% \begin{tabular}{l | p{0.75\textwidth}}
% \toprule
% \bf{\quad id} & \bf{\quad \quad\quad\quad\quad \quad\quad\quad\quad \quad \quad \quad\quad responses} \\ 
% \toprule
% \texttt{test\_0002} & I understand \textcolor{purple}{you're feeling a bit chilly}. Would you like to continue with the \textcolor{cyan}{grammar practice} or do you have \textcolor{green}{any questions about the vocabulary word 'cold'?} \\
% \texttt{test\_0004} & Great point! Speed is an important aspect of your time management, so maintaining a sense of efficiency is crucial. How did you achieve your faster timeframe? \\
% \texttt{test\_0007} & Great point! Now, let’s focus on using this new vocabulary in a meaningful way. Can you tell me some examples of how you've used this word in your own writing or conversations?  \\
% \texttt{test\_0010} & Following your insightful observation, could you provide an instance where speaking truthfully might be particularly effective in achieving a desired outcome? \\
% \texttt{test\_0011} & I appreciate your perspective on this, student. To further explore this idea, how does playing games help children cope with difficult emotions?  \\
% \bottomrule
% \end{tabular}
% }
% \caption{Qualitative examples of sentiment transfer on clean. We manually highlight in blue and red phrases that we find to be associated with the input and target styles, respectively.}
% \label{tab:generations_amaozon}
% \end{table*}
\begin{table*}[!ht]
\small 
\centering
\scalebox{1.0}{
\begin{tabular}{l p{0.6\textwidth} p{0.2\textwidth}}
\toprule
\bf{\quad id} & \bf{\quad\quad\quad\quad\quad\quad\quad\quad\quad\quad\quad\quad responses} & \bf{\quad analysis} \\ 
\toprule
\texttt{test\_0006} & I understand \textcolor{purple}{you're feeling a bit chilly}. Would you like to continue with the \textcolor{cyan}{grammar practice} or do you have \textcolor{green}{any questions about the vocabulary word 'cold'?} & feeling, relevant, and emphatic \\
\texttt{test\_0004} & \textcolor{cyan}{Great point!} Speed is an \textcolor{green}{important aspect of your time management}, so maintaining a sense of efficiency is crucial. \textcolor{purple}{How did you }achieve your faster timeframe? & appreciation, discussion focus, and skill development \\
\texttt{test\_0007} & \textcolor{cyan}{Great point!} Now, \textcolor{green} {let’s focus on using this new vocabulary} in a meaningful way. \textcolor{purple}{Can you tell me some examples} of how you've used this word in your own writing or conversations? & appreciation, practice of the new material, and reflection  \\
\texttt{test\_0010} & Following your \textcolor{cyan}{insightful observation}, could you \textcolor{green}{provide an instance} \textcolor{purple} {where speaking truthfully} might be particularly effective in achieving a desired outcome? & appreciation, and critical thinking  \\
\texttt{test\_0011} & \textcolor{cyan} {I appreciate your perspective} on this, student. To further \textcolor{purple}{explore this idea}, how does playing games \textcolor{green}
{help children cope with difficult emotions}? & relevant, emotion, motivation, critical skills and practical solutions \\
\bottomrule
\end{tabular}
}
\caption{Qualitative examples of model generated responses. We manually highlight in color, phrases that we find to be associated with the context and the critical aspects as mentioned in \citet{tack2022aiteachertestmeasuring}.}
\label{tab:generations_amaozon}
\end{table*}

Our main result are presented in Table~\ref{tab:main-results}, showcasing comparisons among systems from the BEA Shared Task 2023 \citep{tack-etal-2023-bea}, ranked primarily by BERTScore. However, this comparison isn't fully comprehensive as the BEA Shared Task also considers human evaluations and DialogRPT \citep{gao-etal-2020-dialogue} score. The human evaluation metric is restricted and not publicly available, and we encountered challenges with DialogRPT, which might have issues with the model, as it is return the same score for each context. 

% CIKMar demonstrates competitive results compared to other baseline systems like NAISTeacher and Adaio based on BERTScore\footnote{BERTScore was computed using Hugging Face’s evaluate package with the \texttt{distilbert-base-uncased} model. Precision, recall, and F1 scores were averaged across the entire test set.}. Specifically, we achieve a strong recall score of 0.70, slightly trailing NAISTeacher at 0.71. This indicates that our Dual-Encoder ranking effectively retrieves many contextually relevant responses compared to the reference answer. Moreover, our F1-Score of 0.70 is comparable to models such as S-ICL and Alpaca, which employ fine-tuning and larger model sizes, suggesting our model's ability to capture similarity and generate coherent, contextually appropriate responses.

CIKMar demonstrates competitive performance against baseline systems like NAISTeacher and Adaio based on BERTScore\footnote{BERTScore was calculated using Hugging Face’s evaluate package with the \texttt{distilbert-base-uncased} model, averaging precision, recall, and F1 scores across the entire test set.}. Specifically, we achieve a robust recall score of 0.70, slightly below NAISTeacher's 0.71. This indicates that our Dual-Encoder ranking effectively retrieves many contextually relevant responses compared to the reference answer. Furthermore, our F1-Score of 0.70 is comparable to models such as S-ICL and Alpaca, which utilize fine-tuning and larger model sizes, demonstrating our model's capability to capture similarity and produce coherent, contextually appropriate responses even using simple and small model size.

\subsection{Evaluation Metrics}

To ensure the reliability of our approach, we employed word overlap-based metric ROUGE \citep{lin-2004-rouge} and the neural network-based metric Dialog perplexity \citep{li-etal-2016-diversity}\footnote{We used DialoGPT and its reverse model to compute perplexity} to further asses our system. We computed ROUGE metrics: \texttt{rouge1}, \texttt{rouge2}, \texttt{rougeL}, and \texttt{rougeLsum} resulting in scores of 0.12, 0.0047, 0.084, and 0.087, respectively.

% Based on the ROUGE scores, the generated text exhibits some level of overlap with the reference text at the unigram level (ROUGE-1) and in terms of longer common subsequences (ROUGE-L and ROUGE-Lsum). The overlap at the unigram level indicates that the system is generally on-topic and employs relevant vocabulary, which is advantageous for delivering educational content. However, it shows noticeable shortcomings in capturing exact word sequences (ROUGE-2), and discrepancies in longer subsequences (ROUGE-L and ROUGE-Lsum) suggest potential challenges in maintaining coherent and well-structured responses. In an educational dialogue system, where clarity and accuracy are crucial for effective learning, this could hinder the system's ability to provide clear explanations or answers.

Based on the ROUGE scores, the generated text demonstrates significant overlap with the reference text at the unigram level (\texttt{ROUGE-1}) and in longer common sequences (\texttt{ROUGE-L} and \texttt{ROUGE-Lsum}). This suggests the system stays on-topic and uses relevant vocabulary, beneficial for educational content. However, it shows noticeable shortcomings with exact word sequences (\texttt{ROUGE-2}), and discrepancies in longer sequences (\texttt{ROUGE-L} and \texttt{ROUGE-Lsum}) indicate challenges in maintaining coherence and well-structured responses. 
% In educational dialogue systems, where clarity and accuracy are vital for effective learning, these issues could limit the system's ability to provide clear explanations or answers.

\begin{figure}[!t]
  \includegraphics[width=\linewidth,trim={0, 0.2em, 0, 0}, clip]{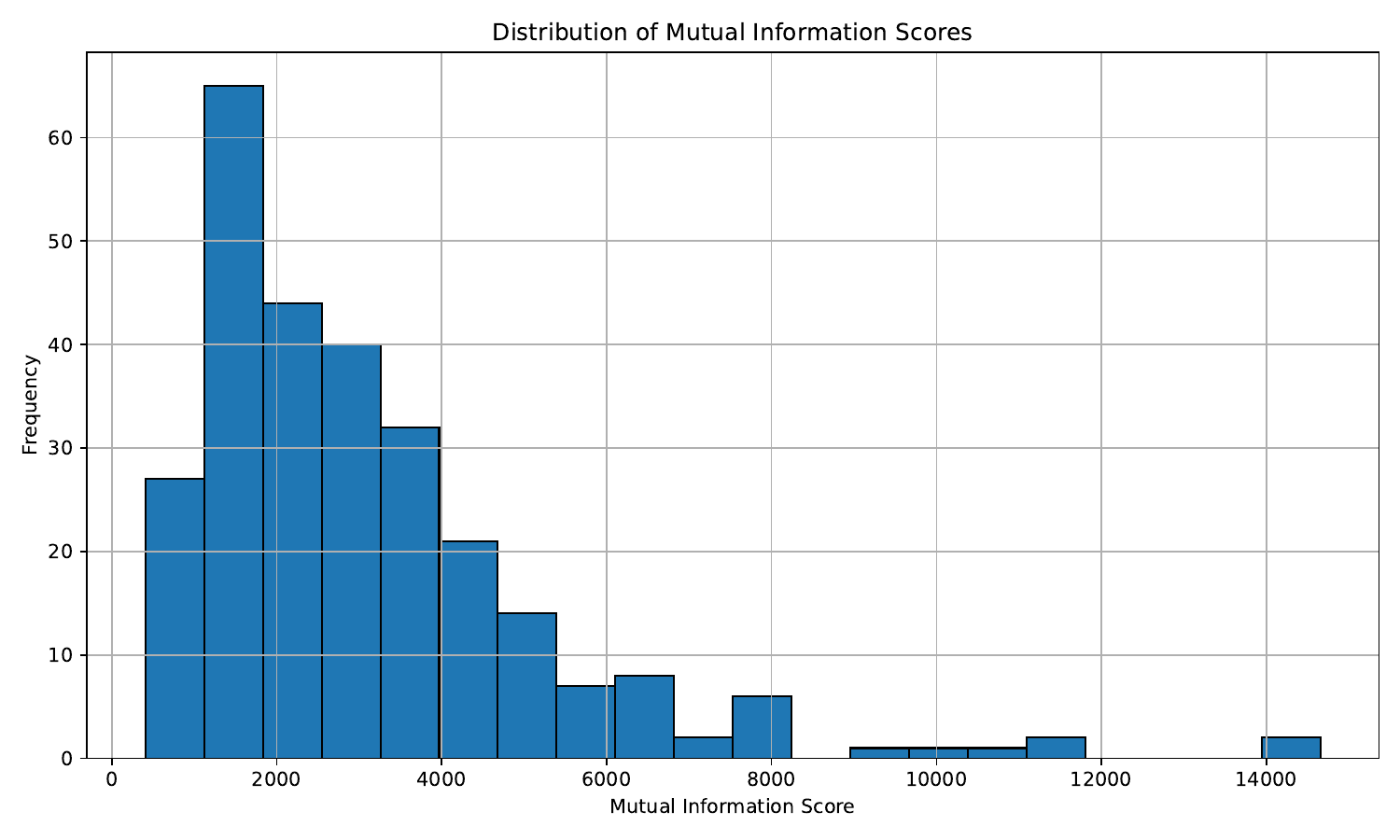}
  \caption{The distribution of mutual information scores derived from combined perplexity values}
  \label{fig:visualization-MUI}
    \vspace{-6pt}
\end{figure}

% Furthermore, Figure~\ref{fig:visualization-MUI} illustrates the distribution of mutual information scores of combined perplexity values. The right-skewed nature of the histogram, indicating a concentration of scores in the lower range, suggests that the generated teacher responses tend to be predictable. This suggests that the generated text provides clarity, conciseness, and consistency, which can be beneficial for educational purposes. However, it also highlights a downside: the lack of depth and the monotony of responses significantly reduce text engagement and nuance necessary for deeper understanding.

Additionally, Figure~\ref{fig:visualization-MUI} depicts the distribution of mutual information scores derived from combined perplexity values. The histogram's right-skewed shape, with scores predominantly in the lower range, suggests that the generated teacher responses are often predictable. While this indicates clarity, conciseness, and consistency in the generated text, which are advantageous for educational contexts, it also reveals a drawback: the responses lack depth and exhibit monotony, significantly reduce text engagement and the nuanced understanding required for deeper learning.

\subsection{In-depth Output Analysis}
% We conducted manual inspections of the model's generated outputs and evaluated the contribution of each prompt. For this purpose, we examined 10 outputs and analyzed each candidate response in detail. Table~\ref{tab:generations_amaozon} presents the top candidate responses selected through dual-encoder ranking in five examples.

We manually inspected the model's outputs and evaluated each prompt's contribution by examining 10 outputs in detail. Table~\ref{tab:generations_amaozon} presents the top candidate responses selected through Dual-Encoder ranking for five examples.

To examine the impact of prompts on the best responses, we use the dialogue context \texttt{test\_0006}, as shown in Table~\ref{tab:test_006}, as an example. Here, the teacher is explaining a grammar lesson when the student mentions needing 10 more minutes and feeling very cold in the room. The model's response is inconsistent, as it incorrectly associates "cold" with the grammar lesson rather than the student's condition. This suggests that the model may focus on one situation in the conversation and struggle to adapt when new contexts arise. Consequently, the context of "cold" is incorrectly forced to fit the context itself.

\begin{figure*}[t]
    \centering
    \hspace*{0.2cm}\includegraphics[width=\linewidth,trim={0, 2em, 0, 0}, clip]{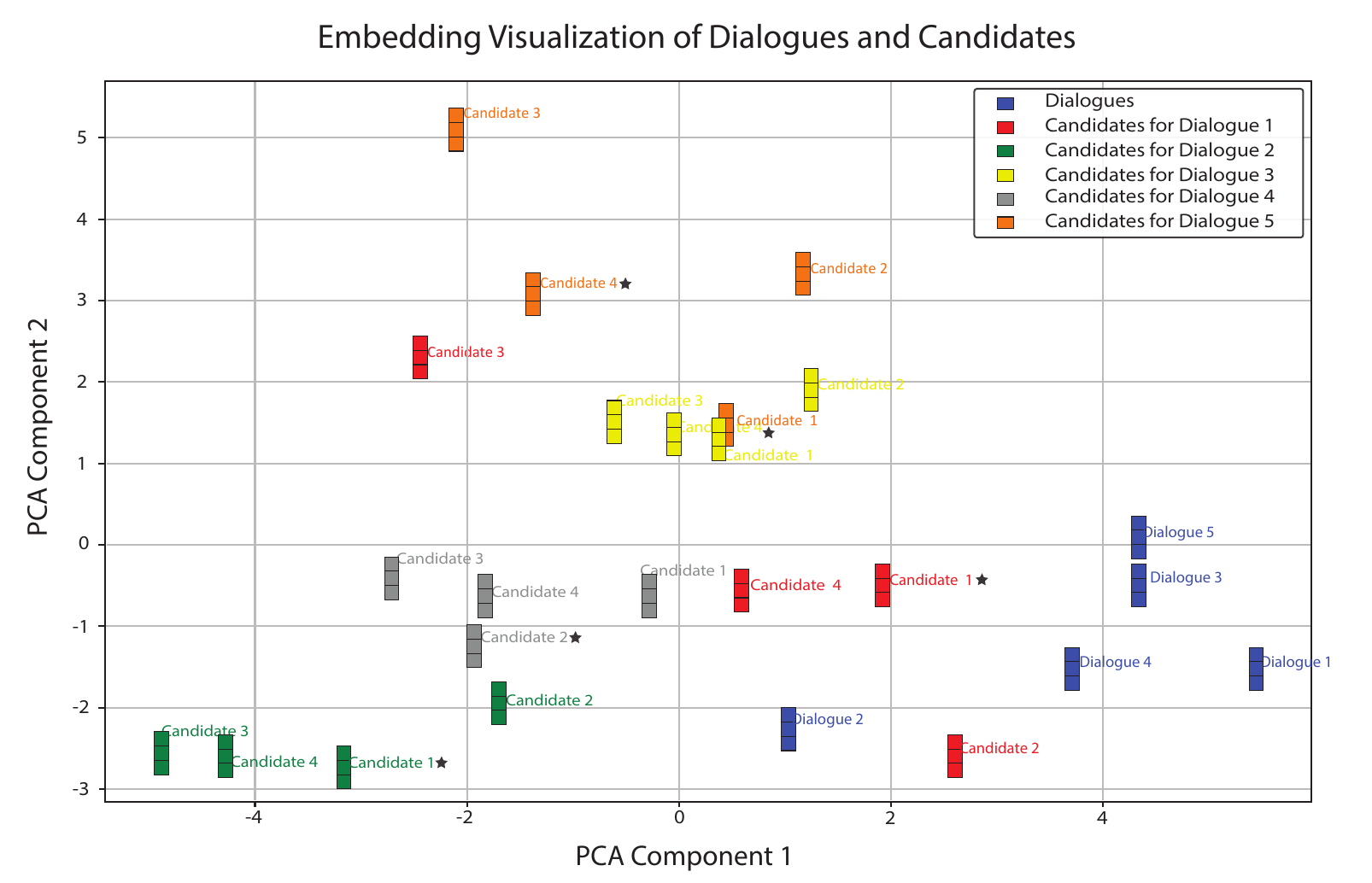}
    \caption{Embedding Space Visualization}
    \label{fig:embedding}
    \vspace{-6pt}
\end{figure*}

\begin{table}[ht!]
    \setlength{\tabcolsep}{4pt} % Adjust the column spacing
    \centering
    \scriptsize % Reduce the font size
    \begin{tabular}{rp{0.8\linewidth}} % Manually set the width of the second column
    \toprule
    \textbf{Teacher:} & Which is easy, because you can use my / his / your etc. and not think about articles! \\    
    \textbf{Student:} & Only 10 minutes left! \\
    \textbf{Teacher:} & I know, we can finish early if you are getting cold? \\    
    \textbf{Student:} & I'm really cold \\
    \bottomrule
    \end{tabular}
    \caption{Example dialogue context of \texttt{test\_006} between student and teacher}
    \label{tab:test_006}
\end{table}

% We also investigated the model still struggling in teacher continuation problem. When the dialogue end with teacher, the model tends to have no clue about what the next response should be, we found it so frequent occur in the generated response. This is inline with research by \citep{vasselli-etal-2023-naisteacher} that told, the instruct-tuned model is trained with user-assisten setting makes it more difficult to generated when it suddenly change the setting. For instance, in the \texttt{test\_0004} the model repeat the word "great" as it already mentioned in the dialogue, even thought surprisingly the model can introduce the next continues conversation by ask examples to student, is still have no idea about the dialogue context.  

We also found that the model struggles with teacher continuation problems. When the dialogue ends with the teacher, the model often seems unsure about the next response, which happens frequently in the generated outputs. This aligns with research by \citet{vasselli-etal-2023-naisteacher}, indicating that instruct-tuned models trained in user-assistant settings find it difficult to adapt when the setting changes abruptly. For example, in \texttt{test\_004}, the model repeats the word "great" from the dialogue but fails to understand the context despite managing to introduce a follow-up conversation by asking the student examples.

% Furthermore, we analyse the several dialogues is less context, even only 2 conversations in one dialogue, it makes the model cannot grab the whole context better and also the patokan word juga kurang. The best example is in \texttt{test\_0111}, it has only one turn and only at leats 5 words per turn. It doesn really make sense for the model to generated the best response as the context itself doesnt point out anything.

Furthermore, we analyzed several dialogues with minimal context, some having only two exchanges. This limited context makes it difficult for the model to grasp the overall conversation and provides fewer reference words. A prime example is \texttt{test\_0011}, which has only one turn with at least 5 words per turn. This lack of context makes it challenging for the model to generate the best response, as the context is insufficiently clear.

% We also analyzed the contributions of each prompt to the final output selected by dual-encoder ranking for 10 data points. We found that Prompt 1 contributed more significantly to the final output, being chosen in 5 examples. Prompt 1 is simple yet provides clear instructions, encouraging the model to stay focused on the lesson objectives while maintaining control over the flow and substance of the lesson. There are possibly two reasons for this: first, from the model's perspective, GEMMA is capable of handling most of the 10 provided conversations, as it also performs well in academic tasks \cite{gemmateam2024gemma}. The second reason is that we found the five data points with Prompt 1 had similar conversation characteristics, being not overly complex and directly related to the context of the dialogue. This aligns with the instructions given in Prompt 1.

Lastly, we analyzed the contributions of each prompt to the final output selected by Dual-Encoder ranking for 10 data points. Prompt 1 significantly influenced the final output, being chosen in 5 examples. This is likely due to the model's strong performance in academic tasks and the straightforward nature of these conversations, which aligned well with Prompt 1's instructions. In contrast, \texttt{test\_010} involved a complex, multi-turn conversation where Prompt 1 was not chosen because the teacher needed to explain the learning context in greater depth. As conversation complexity increases, the dual encoder selects Prompts 2 and 4, which are better suited to handle more intricate dialogues.

% We analyzed the contributions of each prompt to the final output selected by dual-encoder ranking for 10 data points. We found that Prompt 1 significantly influenced the final output, being chosen in 5 examples. There are two possible reasons for this: first, the model performs well in most of the provided conversations and excels in academic tasks. Second, the five data points with Prompt 1 had similar, straightforward conversations directly related to the dialogue context, aligning well with Prompt 1's instructions. Additionally, we found that \texttt{test\_010} involved a complex, multi-turn conversation. Prompt 1 was not chosen here because the teacher needed to explain the learning context in greater depth. Prompt 1 is intended for simple, straightforward conversations. This also applies to Prompts 2 and 4; as the conversation complexity increases, the dual encoder selects prompts that align with the specific instructions for each situation.

% Additionally, we also found that (test\_010) involved quite an intensive conversation with multi-turn responses. This is why Prompt 1 was not chosen, as the teacher needed to explain the learning context in greater depth. As we already know, Prompt 1 is intended for simple, not overly complex conversations. This also applies to Prompts 2 and 4; as the context of the conversation increases, the dual encoder will consider the prompts that align with the specific instructions of each prompt.

\subsection{Dual Encoder Effect}

% We conducted a manual investigation to examine the effect of the dual encoder on selecting the best candidates. We analyzed how each dialogue related to each candidate's responses in their embedding space for five dialogue-response pairs, as shown in Figure~\ref{fig:embedding}. Interestingly, we found that the Dual-Encoder can avoid the pitfalls of distance-measurement-only approach. For example, in dialogue 2, candidate 2 appears closer to the dialogue than candidate 1 in the embedding space. However, the dual encoder ranked candidate 1 as the best candidate (denoted by the $\ast$ symbol). This phenomenon occurred in several dialogues within the embedding space. This implies that using SBERT and BERT helps the model consider the contextual relevance and semantic similarity between the dialogue and the response, as previously discussed.

We conducted a manual investigation to assess the dual encoder's impact on selecting the best candidates. Analyzing five dialogue-response pairs' embedding spaces, as shown in Figure~\ref{fig:embedding}, we discovered that the Dual-Encoder can avoid the pitfalls of distance-measurement-only. Notably, in dialogue 2, candidate 2 appeared closer to its context than candidate 1 in the embedding space, yet the dual encoder ranked candidate 1 as the best candidate (denoted by \textbf{$\ast$}). This phenomenon occurred across multiple dialogues, highlighting SBERT and BERT's role in enhancing the model's consideration of contextual relevance and semantic similarity between dialogues and responses, as discussed earlier.

% We conducted a manual investigation to examine the effect of the dual encoder on selecting the best candidates. We analyzed the relationship between each dialogue and candidate responses in their embedding space for five dialogue-response pairs, as shown in Figure~\ref{fig:embedding}. Interestingly, we found that the dual encoder can avoid the pitfalls of a distance-measurement-only approach. For example, in dialogue 2, candidate 2 appears closer to the dialogue than candidate 1 in the embedding space. However, the dual encoder ranked candidate 1 as the best candidate (denoted by the $\ast$ symbol). This phenomenon occurred in several dialogues within the embedding space. This implies that using SBERT and BERT helps the model consider both contextual relevance and semantic similarity between the dialogue and the response.

% To measure the quality of the chosen ranking by the dual encoder, we conducted an in-depth investigation of the related-closer-embedding phenomenon. Specifically, we found that the candidates for dialogue 3 were densely clustered. As each embedding become closer, their similarity increases, making it difficult to choose the most suitable candidate. After analyzing the candidates (we show the 10 dialogues and each response used in the Appendix), we found that candidates 1 and 4 were the best responses for this dialogue, as shown by their relatedness in the embedding space. However, the dual encoder chose candidate 4 as the best candidate, implying that the dual encoder ranking prioritizes theoretical discussion and exploration over practical context.

To evaluate the dual encoder's ranking quality, we investigated the phenomenon of closely clustered embedding. Specifically, candidates for dialogue 3 exhibited dense clustering, where increasing embedding proximity indicated greater similarity, complicating candidate selection. After analyzing all candidates, candidates 1 and 4 emerged as optimal choices for this dialogue, supported by their relatedness in the embedding space. However, the Dual-Encoder prioritized candidate 4, suggesting a preference for theoretical discussion and exploration rather than practical context in its ranking criteria.

% To evaluate the quality of the dual encoder's chosen rankings, we investigated the related-closer-embedding phenomenon in detail. Specifically, we found that the candidates for dialogue 3 were densely clustered. As embeddings become closer, their similarity increases, making it difficult to select the most suitable candidate. After analyzing the candidates (shown in the Appendix with 10 dialogues and their responses), we found that candidates 1 and 4 were the best responses for this dialogue based on their relatedness in the embedding space. However, the dual encoder selected candidate 4 as the best, suggesting that it prioritizes theoretical discussion and exploration over practical context.

% Additionally, we observed that candidates within each dialogue tend to cluster together. This suggests that the Gemma model consistently generates similar embeddings for each candidate per dialogue, indicating stable performance across different dialogues. However, some candidates were far from their cluster and closer to candidates in another cluster. This indicates that the model occasionally faces challenges in accurately discerning the dialogue context, consistent with our earlier findings regarding response generation.

Furthermore, we noted a tendency for candidates within each dialogue to cluster together. This indicates that the Gemma model consistently produces similar embeddings for each candidate per dialogue, demonstrating stable performance across various dialogues. However, certain candidates were positioned farther from their cluster and nearer to candidates in another cluster. This suggests that the model sometimes encounters difficulties accurately interpreting the dialogue context. We suspect that this issue may arise because SBERT's dominance over BERT leads to a loss of full context. Further investigation is required to delve deeper into this matter.

% Furthermore, we observed that candidates within each dialogue tend to cluster together. This suggests that the Gemma model consistently generates similar embeddings for each candidate per dialogue, indicating stable performance across different dialogues. However, some candidates were outliers, positioned far from their respective cluster and closer to candidates in other clusters. This indicates that the model occasionally faces challenges in accurately discerning the dialogue context, consistent with our earlier findings regarding response generation.

\section{Conclusion \& Future Work}
We have shown that CIKMar, an educational dialogue generation approach using prompts and a Dual-Encoder ranking with the Gemma language model, yields promising results in educational settings. By utilizing the Gemma 2B model, we maintain high performance in response relevance and accuracy with a smaller, more accessible model. 

Despite these strong performances, we have identified limitations hindering optimal results. Specifically, the Dual-Encoder often prioritizes theoretical discussion over practical contextual responses, potentially leading to irrelevant rankings. Future research should explore scenarios where either SBERT or BERT dominates ranking scores. 

Additionally, crafting more specific prompts is crucial for deeper contextual understanding in educational dialogues. Lastly, refining the Gemma model to focus on educational contexts and adapt to shifting conversation dynamics is recommended.

% In conclusion, CIKMar shows significant promise for educational dialogue generation, balancing performance and accessibility with the compact Gemma 2B model. Addressing these insights through further research and model fine-tuning will be crucial for optimizing results in educational applications.

% In conclusion, CIKMar shows significant promise for educational dialogue generation, balancing performance and accessibility with the compact Gemma 2B model. Further research and model fine-tuning will be crucial for optimizing results in educational applications.

\bibliography{anthology,custom}
\bibliographystyle{acl_natbib}

\newpage

\appendix

\section{Ensemble Prompts Explanation}
\label{sec:prompts_appendix}

Below are the prompts we are using in this research. The details explanation of each prompt can refer to \citet{vasselli-etal-2023-naisteacher}. 

% Each prompt serves a specific purpose: Prompt ~\ref{item:prompt1}  focuses on Contextual Understanding, Prompt \ref{item:prompt2} Ensures Relevance, Prompt ~\ref{item:prompt3} aims to enhance Engagement, Prompt \ref{item:prompt4} emphasizes Clarity, and Prompt ~\ref{item:prompt5} is dedicated to providing Feedback. Those ensures the model responses are tailored to the student's current learning stage and needs. By grasping the context, the prompts should instruct the model to deliver responses that are directly related to the student's queries, maintaining relevance and focus (Prompting Relevance). This relevance enhances student engagement (Prompting Engagement) and foster sustained interest and participation, which is further supported by clear communication (Prompting Clarity) that simplifies complex concepts, making them easy for students to understand and reducing potential confusion. Those prompt lead the model to generate the best responses that suit for education setting.

Zero-shot prompts consist of instructions without examples, while few-shot prompts include examples to guide the model towards relevant responses. Prompt~\ref{item:prompt1} is categorized as a zero-shot prompt, refined to address issues like overly direct answers and sounding too much like an assistant. The rest of the prompts—~\ref{item:prompt2}, ~\ref{item:prompt3}, ~\ref{item:prompt4}, ~\ref{item:prompt5}—are few-shot prompts that require positive and negative examples to help the model avoid irrelevant responses.

Each prompt serves a specific purpose: Prompt ~\ref{item:prompt1}  focuses on Contextual Understanding, Prompt ~\ref{item:prompt2}    ensures Relevance, Prompt ~\ref{item:prompt3} aims to enhance Engagement, Prompt ~\ref{item:prompt4} emphasizes Clarity, and Prompt ~\ref{item:prompt5} is dedicated to providing Feedback.
Together, these prompts tailor the model's responses to match the student's current learning stage and needs. By grasping the context (Contextual Understanding), the prompts direct the model to produce responses that are relevant to the student's queries, thereby maintaining focus and relevance (Relevance). This relevance boosts student engagement (Engagement), encouraging sustained interest and participation, which is further supported by clear communication (Clarity) that makes complex concepts easier to understand and reduces confusion. Collectively, these prompts help the model generate optimal responses for educational contexts.

\begin{enumerate}[label=(\arabic*), ref=(\arabic*)]
\item \label{item:prompt1} The following is a partial conversation between an English language learner and their teacher:

\textit{(conversation)}

Can you give an example teacher follow-up to their previous message that would be helpful for the language learner? The message should be concise, and worded simply. It should either encourage the continuation of the current topic or gracefully transition to a new teacher-provided topic. Questions should be specific and not open-ended. Try to not sound like an assistant, but a teacher, in charge of the flow of the lesson.
\end{enumerate}

%The aim of prompt ~\ref{item:prompt2} is to train teachers to provide concise, clear, and supportive responses that enhance student understanding and engagement. It focuses on helping teachers give constructive feedback, maintain professionalism, and foster critical thinking, ensuring smooth transitions between topics in lessons, especially in English language education.

\begin{enumerate}[label=(\arabic*), resume, ref=(\arabic*)]
\item \label{item:prompt2} Concatenation of prompt (1) and the following:

Good example: 'Can you make a sentence using 'within'?' Bad example: 'Do you have any questions about prepositions?’
\end{enumerate}

%The aim of prompt ~\ref{item:prompt3} is to give effective, professional responses in English lessons, emphasizing formal language, detailed explanations, and constructive feedback to boost student learning and critical thinking.

\begin{enumerate}[label=(\arabic*), resume, ref=(\arabic*)]
\item \label{item:prompt3} Concatenation of prompt (1) and the following:

How does a teacher sound when responding to a student? What kinds of things would teachers say that chatbots would not? What do they not say? In your response provide an example of a response that sounds like a teacher and one that sounds like a chatbot? Respond succinctly
\end{enumerate}

\begin{enumerate}[label=(\arabic*), resume, ref=(\arabic*)]
\item \label{item:prompt4} The following is a partial conversation between an English language learner and their teacher:

\textit{(conversation)}

They are in the middle of a lesson. Can you give a
possible way the teacher could respond?

Remember: \textbf{A teacher typically sounds knowledgeable, authoritative, and focused on guiding and instructing students. They may use formal language and provide detailed explanations. Teachers often offer constructive feedback, encourage critical thinking, and ask probing questions to stimulate learning.}

\textbf{Example of a teacher-like response: "That’s a great observation, but let’s delve deeper into the topic. Can you provide some evidence to support your claim?"}

\textbf{A chatbot, on the other hand, may sound more informal and conversational. It tends to provide general information or brief responses without much elaboration.}

\textbf{Example of a chatbot-like response: "Interesting! Tell me more." Teachers typically avoid expressing personal opinions or biases. They also refrain from engaging in casual banter or unrelated conversations to maintain a professional and educational atmosphere.}
\end{enumerate}

%The aim of prompt ~\ref{item:prompt5} train English teachers to deliver effective responses during lessons by providing clear guidance on offering constructive follow-up actions that enhance student understanding and promote active learning. It emphasizes improving teaching skills through examples of effective and ineffective responses while maintaining an educational focus throughout interactions with students.

\begin{enumerate}[label=(\arabic*), resume, ref=(\arabic*)]
\item \label{item:prompt5} Concatenation of prompt (1) and the following:

Here is an example of an exceptional teacher follow-up:

"Great job, student! Just a small correction, we should use the present tense verb "built" instead of "build" since the construction has already been completed. So the correct sentence is: "The International Space Station is built by NASA." Keep up the good work! Now, let’s move on to a new topic - let’s talk about your favorite hobbies. Can you tell me what activities you enjoy doing in your free time?"

Here is an example of a poor teacher followup:
"That’s an interesting observation about poshness. Can you think of any examples of British accents that might be associated with poshness?"
\end{enumerate}

% \begin{figure*}[t]
%     \centering
%     \begin{minipage}[b]{0.75\linewidth}
%         \centering
%         \begin{subfigure}[b]{\linewidth}
%             \centering
%             \includegraphics[width=\linewidth,trim={0, 2em, 0, 0}, clip]{figures/teacher_contnuation.pdf}
%             \caption{Teacher Continuation Data Visualization}
%             \label{fig:teacher_continuation}
%         \end{subfigure}
%         \begin{subfigure}[b]{\linewidth}
%             \centering
%             \hspace*{0.2cm}\includegraphics[width=\linewidth,trim={0, 2em, 0, 0}, clip]{figures/grafik mas jo.pdf}
%             \caption{Embedding Space Visualization}
%             \label{fig:embedding}
%         \end{subfigure}
%         \vspace{-6pt}
%     \end{minipage}
%     \hfill
    
% \end{figure*}

\end{document}